\documentclass[letterpaper, 10 pt, conference]{ieeeconf}  
\IEEEoverridecommandlockouts                              
\usepackage{graphics} 
\usepackage{graphicx}
\usepackage{epstopdf}
\usepackage{hyperref}
\usepackage{epsfig} 
\usepackage{times} 
\usepackage{soul, color, xcolor}
\usepackage{cite}
\usepackage{amsmath,amssymb,amsfonts,bm}
\usepackage{makecell}
\usepackage{algorithm}
\usepackage{algorithmic}
\usepackage{booktabs}
\usepackage{multirow}
\usepackage{textcomp}
\usepackage{listings} 
\title{\LARGE \bf
Few-Shot Point Cloud Semantic Segmentation via\\ Contrastive Self-Supervision and Multi-Resolution Attention
}
\author{Jiahui Wang, Haiyue Zhu$^\dagger$, Haoren Guo, Abdullah Al Mamun, Cheng Xiang and Tong Heng Lee
\thanks{$\dagger$ Corresponding author: zhu\_haiyue@simtech.a-star.edu.sg}
\thanks{J. Wang, H. Guo, A. Mamun, C. Xiang and T. H. Lee are with College of Design and Engineering, Electrical and Computer Engineering, National University of Singapore, 117582, Singapore. }
\thanks{H. Zhu is with Singapore Institute of Manufacturing Technology (SIMTech), Agency for Science, Technology and Research (A*STAR), 2 Fusionopolis Way, Singapore 138634, Republic of Singapore.
} \\%
}

\begin{document}
\maketitle
\thispagestyle{empty}
\pagestyle{empty}
\begin{abstract}

This paper presents an effective few-shot point cloud semantic segmentation approach for real-world applications. Existing few-shot segmentation methods on point cloud heavily rely on the fully-supervised pretrain with large annotated datasets, which causes the learned feature extraction bias to those pretrained classes. However, as the purpose of few-shot learning is to handle unknown/unseen classes, such class-specific feature extraction in pretrain is not ideal to generalize into new classes for few-shot learning. Moreover, point cloud datasets hardly have a large number of classes due to the annotation difficulty. To address these issues, we propose a contrastive self-supervision framework for few-shot learning pretrain, which aims to eliminate the feature extraction bias through class-agnostic contrastive supervision. Specifically, we implement a novel contrastive learning approach with a learnable augmentor for a 3D point cloud to achieve point-wise differentiation, so that to enhance the pretrain with managed overfitting through the self-supervision. Furthermore, we develop a multi-resolution attention module using both the nearest and farthest points to extract the local and global point information more effectively, and a center-concentrated multi-prototype is adopted to mitigate the intra-class sparsity. Comprehensive experiments are conducted to evaluate the proposed approach, which shows our approach achieves state-of-the-art performance. Moreover, a case study on practical CAM/CAD segmentation is presented to demonstrate the effectiveness of our approach for real-world applications.

\end{abstract}

\section{INTRODUCTION}
Point cloud semantic segmentation is a significant task in computer vision, which benefits many practical applications such as intelligent manufacturing, autonomous driving, robot manipulation, etc. Due to its complexity and unstructured property, point cloud segmentation learning generally requires more resources including annotated datasets, computation hardware, etc. Although existing fully-supervised methods have demonstrated their capability ~\cite{qi_pointnet_2017,qi_pointnet_2017-1,thomas_kpconv_2019,wang_dynamic_2019,zhao_point_2021,lai_stratified_2022}, one problem with these methods is that they heavily rely on large-scale annotated datasets, which are unpractical or costly to obtain for many real-world applications. In addition, models trained under fully-supervised learning has relatively poor generalization capability to novel classes. As many practical applications demand the learning capability in a data-efficient manner, few-shot point cloud semantic segmentation~\cite{zhao_few-shot_2021,li_few-shot_2021,ren-few-2022-icra} gains more and more popularity due to its effectiveness and practicability.

Meta-learning and metric-learning are commonly used branches of few-shot learning, which mostly follow pretrain, train and test paradigm~\cite{wang_generalizing_2020,tian_generalized_2022}. By embracing few-shot learning, point cloud segmentation models generally have the abilities to learn more effectively from a relatively small annotated dataset with better generalization performance. WPS-Net~\cite{wang2020few} proposed a learnable coherent point transformer for point cloud segmentation that can predict a smooth geometric transformation field to morph the template 3D shape towards the input shape. The multi-view comparison method is also developed for few-shot point cloud semantic segmentation~\cite{chen-compositional}. Recently, attMPTI~\cite{zhao_few-shot_2021} proposed a notable multi-prototype transductive method that captures the geometric dependencies and semantic correlations between points. Our previous work~{\cite{wang2022cam}} adopts T-Net~{\cite{qi_pointnet_2017-1}} to enhance the model comprehension ability with a center loss. However, its performance is still limited in complex scenarios.

Traditionally, few-shot point cloud segmentation is usually pretrained using fully-supervised learning, where the reinforced feature projection is highly coupled with those pretrained classes leading to class-specific feature bias. As a result, such biased feature extraction might be harmful to the few-shot generalization as its purpose is to handle unseen/unknown classes. Moreover, fully-supervised pretrain generally requires a large-scale annotated dataset in a close domain to the few-shot tasks, which may not be practically feasible for many real-world tasks. 
To address such weaknesses, we adopt the contrastive self-supervision technique~\cite{guo2022masked} for few-shot point cloud segmentation to eliminate such feature extraction bias, so that the learned encoder projection can be more universal to extract better and more general features for further few-shot learning. Moreover, as our approach is self-supervised, it largely removes the burden to prepare an annotated dataset for the pretrain purpose. 

In this work, we propose learnable-augmentor-based contrastive learning to achieve the effective self-supervised pretrain, where the learning objective is to differentiate the individual points in two alternative views. Unlike 2D image augmentation, point cloud augmentation process data in an unstructured and out-of-order format. Consider these characteristics, we integrate a trainable augmentor from PointAugment~\cite{li_pointaugment_2020} in our contrastive learning which can automatically optimize the augmentation to exploit better unstructured variation and deformation in both shape and point levels. As a result, the self-supervised feature extraction is reinforced in the differentiation capability on the point level without class-specific guidance, which benefits the following few-shot adaptation.

Moreover, we propose a Multi-Resolution Attention (MRA) module for the few-shot learning to better capture both the local and global information. Although the attention mechanisms~\cite{vaswani_attention_2017, guo_pct_2021} are widely used in point cloud tasks, most approaches directly use attention to all points, which includes excessive distractive information from the redundant background and affects the performance. This problem is especially more serious when segmenting small objects, where the vast background points can easily overwhelm the foreground feature. Based on this finding, we design a MRA block that only conducts attention to local nearest neighbours and global farthest points. Due to the selective point attention, our approach also saves the computation memory and achieves better performance.

Finally, the contribution of this work can be summarized as,
\begin{itemize}
    \item We propose a few-shot point cloud semantic segmentation framework with contrastive self-supervised pretrain, which can not only mitigate the influence of class-specific bias but also significantly reduce the requirement for annotated large-scale pretrain datasets.
    \item A computation-efficient MRA block is proposed, which is able to conduct attention more effectively and reduce the influence of distractive background information.
    \item We propose a centering-based multi-prototype generation process that can decrease the intra-class distance and enlarge the inter-class distance, which ensures the dense distribution within the class and discrimination between classes.
\end{itemize}


\begin{figure*}[htbp]
\centering
\includegraphics[width=17cm]{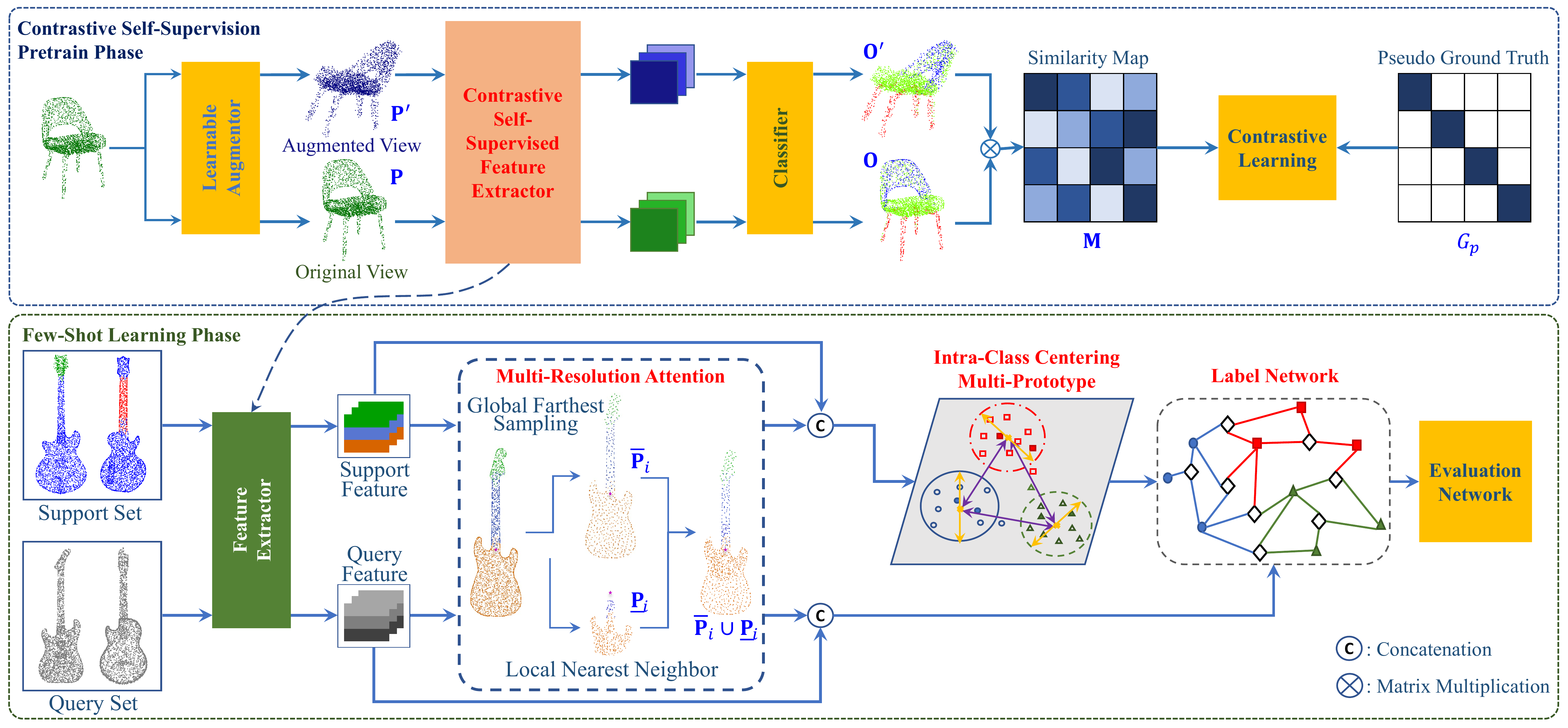}
\caption{Our framework includes pretrain and few-shot learning phases. In the pretrain phase, a trainable augmentor is employed to generate the augmented view of input point cloud, where a point-wise contrastive loss is designed to self-supervise the encoder. After the pretrain phase, only the encoder will be used. In the few-shot learning phase, the encoder maps input point clouds to feature space, and a Multi-Resolution Attention (MRA) block extracts the attention feature and concatenates it with the input. With the support feature, the centering-based multi-prototype generation block extracts prototypes and ensures the dense distribution in feature space. The label network then takes prototypes and query features as input and makes graph-based label propagation. Lastly, the loss function is calculated in the evaluation block with the ground-truth label of the query set.}
\label{fig:netstructure}
\end{figure*}

\section{RELATED WORK}
\subsection{Point Cloud Semantic Segmentation}
Point cloud semantic segmentation can be treated as a point-wise classification problem, which needs to consider point-to-point difference and topology distribution. PointNet~\cite{qi_pointnet_2017} firstly expresses an end-to-end deep neural network can directly deal with raw point clouds and obtain reasonable performance rather than transforming point clouds to voxel grids or multi-view images. However, PointNet lacks the ability to extract local geometric features. PointMLP~\cite{ma2022rethinking} extracts local features using residual multi-layer perceptron (MLP) blocks and a geometric affine module. While DGCNN~\cite{wang_dynamic_2019} proposed Edgeconv block to enhance the ability of capturing local features. With the increasing popularity of the attention mechanism, Stratified Transformer~\cite{lai_stratified_2022} addresses this problem with an attention-based block. In our work, we propose an MRA block that can not only extract local nearest-neighbour features but also reserve the global key-shape features. 

\subsection{Few-Shot Learning}

Few-shot learning is a research area to deal with sample scarcity. It aims to train a model which is capable of generalizing to new classes without a large number of labeled data~\cite{yiting2021few}. Several approaches have been proposed to address the few-shot issue, for example, transfer learning~\cite{weiss2016survey} transfers experience from the source task with abundant labeled data to the target tasks, where available data is scarce. Meta-learning~\cite{li_deep_2021,wang_hybrid_2019} trains a meta-learner that can extract meta-knowledge across tasks. Among these methods, metric-learning is noticeable because of its simplicity and effectiveness. Metric-learning-based approaches usually have the paradigm that it computes the representation of each support sample to propagate the information to the query feature and assigns each query instance to the support class with the largest similarity to itself. Few-shot learning is also capable to collaborate with other learning algorithm such as incremental learning~\cite{li2022incremental} to achieve better generalization ability. In the point cloud few-shot domain, attMPTI~\cite{zhao_few-shot_2021} represents each class with multiple prototypes in the feature space and apply a non-parametric method to predict the classes for query instance based on the support feature. In our work, we propose a contrastive self-supervision to enhance the feature extraction ability for the few-shot learning through the pretrain phase, and a MRA module with a metric-learning-based prototype and label propagation is proposed to facilitate the few-shot adaptation.

\section{Methodology}

\subsection{Problem Formulation}
According to the previous works~\cite{zhao_few-shot_2021}, the few-shot point cloud semantic segmentation to address in this work is a $N$-way $K$-shot segmentation learning problem. Given a support set $\mathcal{S}=\big\{\{\mathbf{P}_{s}^{n,k}, \bm{Y}_{s}^{n,k}\}_{k=1}^{K}\big\}_{n=1}^{N}$ with $N$ classes in total to segment, each $n$-th class contains a number of $K$ support point clouds $\mathbf{P}_{s}$ with their corresponding labels $\bm{Y}_{s}$, our learning task is to obtain an optimal network $\bm{F}_{\theta^{*}}(\cdot)$ through $\mathcal{S}$ that can perform the segmentation task on the query set  $\mathcal{Q}=\{\mathbf{P}_{q}, \bm{Y}_{q}\}_{i}$, i.e.,
\begin{equation}
    \begin{aligned}
    \bm{Y}_{q}^{'}&=\bm{F}_{\theta}(\bm{P}_{q}| \mathcal{S}),\ \bm{P}_{q} \in \mathcal{Q},\\
    \theta^{*}&=\underset{\theta}{\operatorname{argmin}}\ \sum_{\mathbf{P}_{q}\in \mathcal{Q}} \mathcal{L}\left(\bm{Y}_{q},\bm{Y}_{q}^{'} \right),    
    \end{aligned}
\end{equation}
where $\mathbf{P}_{s} \in \mathbb{R}^{M\times f}$ contains $M$ points with $f$ channels of features including $xyz$ coordinates, RGB channels, etc., $\bm{Y}_{q}$ and $\bm{Y}_{q}^{'} \in \mathbb{R}^{M\times 1}$, $\mathcal{L}$ is the loss function, 

Following the common practices, the learning procedures for few-shot point cloud segmentation is generally divided into two phases, i.e., pretrain phase and few-shot learning phase. The point cloud dataset $\mathcal{D}=\mathcal{D}_{train}\cup\mathcal{D}_{test}$ contains two non-overlapping subsets $\mathcal{D}_{train}$ and $\mathcal{D}_{test}$ with class sets $\mathcal{Z}_{train}$ and $\mathcal{Z}_{test}$, respectively, where $\mathcal{Z}_{train} \cap \mathcal{Z}_{test} = \varnothing$. The pretrain phase aims to obtain a better base point cloud feature extractor so that it can facilitate the following few-shot learning. Usually, fully-supervised learning is adopted in pretrain to obtain a better initialization of the feature extractor. The few-shot learning phase employs the standard episode learning for the training and testing, where each episode consists of $N\times K$ support point clouds and $N$ query point clouds to form a $N$-way $K$-shot episode. In both the pretrain phase and  training part of the few-shot learning phase, the testing dataset $\mathcal{D}_{test}$ with class set $\mathcal{Z}_{test}$ is not accessible for the training supervision.

\subsection{Framework Overview}

Fig.~\ref{fig:netstructure} illustrates our proposed framework for few-shot point cloud semantic segmentation. In this work, our key idea is to improve the feature extraction capability through the pretrain phase to obtain a better extractor initialization for few-shot learning,  where a contrastive self-supervision learning approach is proposed to serve this purpose. The self-supervised encoder projection $E_{\alpha}(\cdot)$ is supposed to be less biased to pretrained classes, and thus extract better and more general features for further few-shot learning. 

In the few-shot learning phase, our pipeline contains five parts as illustrated in Fig.~\ref{fig:netstructure}. Firstly, the pretrained DGCNN-based encoder $\bm{E}_{\alpha}$ projects both the query and support point clouds into the feature space, denoted as
\begin{equation}
\begin{aligned}
\mathbf{X}_{s/q}&=\bm{E}_{\alpha}(\mathbf{P}_{s/q}).
\end{aligned}
\end{equation}
In this work, we propose an MRA block to better extract both the local and
global information with computational efficiency, denoted as,
\begin{equation}
\begin{aligned}
\mathbf{X}_{s/q}^{'}&=\bm{MRA}_{\alpha}(\mathbf{X}_{s/q})\oplus\mathbf{X}_{s/q}.
\end{aligned}
\end{equation}
which takes the residual structure that concatenates the original feature with the output of MRA, and $\oplus$ denotes the concatenation operation. Next, we sample multi-prototype for each class in the MRA feature space to better represent their complex data distribution. Since the multiple prototypes representing the same class are unregularized, which may have large intra-class distances, we use a centering block to ensure the prototypes have dense distribution in feature space. With prototypes of support samples learned, the prediction of query samples is finished in the label network, which takes $\mathbf{X}_{q}^{\prime}$ and multi-prototype of $\mathbf{X}_{s}^{\prime}$ as input. With label propagation, the network delivers the prediction of each point. Finally, the evaluation network calculates the whole loss function $\mathcal{L}$ of our network and updates the network parameters by gradient descent.

\subsection{Contrastive Self-Supervision with Learnable Augmentor}

We propose a learnable-augmentor-based contrastive self-supervision technique in the pretrain phase to eliminate the feature bias due to the traditional direct class-specific fully-supervised pretraining.  Uniquely, our self-supervision employs a trainable point cloud augmentor to obtain two alternative views for more effective contrastive learning. Inspired by~\cite{li_pointaugment_2020}, the augmentor first projects the original point cloud ($xyz$) $\mathbf{P} \in \mathbb{R}^{M\times 3}$ into a higher-dimension feature space and obtain $\Tilde{\mathbf{P}} \in \mathbb{R}^{M\times C_a}$, where $C_a\gg3$ denotes the feature dimension of the augmentor. Both point-level and shape-level augmentations are considered. The shape-level augmentation trains an MLP network to produce a transformation matrix $\mathbf{T}_{s}$. The point-level augmentation is a trainable MLP producing a displacement noise $\mathbf{N}_{p} \in \mathbb{R}^{M \times 3}$. For shape-level augmentation, we firstly use max pooling on $\Tilde{\mathbf{P}}$ to obtain a feature vector $F_{s} \in \mathbb{R}^{1 \times C_a}$, merge it with a Gaussian distribution noise and fed to the MLP to get  $\mathbf{T}_{s}$. For point-level augmentation, we directly fed our network $N$ copies of $F_{s}$ together with $\Tilde{\mathbf{P}}$ to obtain $\mathbf{N}_{p}$. Lastly, our augmented point cloud for contrastive learning can be denoted as,
\begin{equation}
    \mathbf{P}^{\prime}=\mathbf{P} \otimes \mathbf{T}_{s} + \mathbf{N}_{p},
\end{equation}
where $\otimes$ refers to matrix multiplication. The augmentor also have multiple non-trainable random augmentation layer such as translation and jittering to enhance the augmentation.

Different from the standard contrastive learning in the instance level, our proposed contrastive learning for point cloud is to differentiate in the point level. Specifically, the two alternative views $\{\mathbf{P},\mathbf{P}^{\prime}\}\in \mathbb{R}^{M\times f}$ are fed into the encoder $E_\alpha$ with the classification head to get two segmentation outputs $\{\mathbf{O},\mathbf{O}^{\prime}\}\in \mathbb{R}^{M\times N_{c}}$, where $N_{c}$ is the number of total classes in the pretrain. The contrastive loss is 
formulated to differentiate the same points from $\mathbf{O}$ to $\mathbf{O}^{'}$ as,
\begin{equation}
    \mathcal{L}_{c}=
    \underset{\bm{o}_i \in \mathbf{O}}
    {\mathbb{E}}\left[-\log \left(\frac{e^{\bm{o}_i (\bm{o}_i^{\prime})^{\top}}}{\sum_{j=1}^{M} e^{\bm{o}_i (\bm{o}_j^{\prime})^{\top}}}\right)\Big|\bm{o}_{j}^{'} \in \mathbf{O}^{'}\right],
\end{equation}
where $o_i$ and $o_i^{\prime}$ are the $i$-th point features in $\mathbf{O}$ and $\mathbf{O}^{\prime}$, respectively.

\subsection{Multi-Resolution Attention \& Multi-Prototype Module}

The multi-resolution attention module is proposed to better enhance both the local and global feature extraction, which with the advantages of computational efficiency and effectiveness. We argue that for the local information, the nearest points are of most critical interest as they are most likely to share the same class. For the global information, key points should be sampled globally to represent the overall scene/body instead of with restrictions in a certain level ``bigger" window. Therefore, our key idea is to make use of both the nearest and farthest points as the core concept of multi-resolution attention.

Specifically, consider one query point $P_{i} \in \mathbb{R}^{1 \times f}$ in a query point cloud $\mathbf{P} \in \mathbb{R}^{M \times f}$, our approach finds its K-Nearest Neighbours (KNN) set, denoted as
\begin{equation}
    \begin{aligned}
    \underline{\mathbf{P}}_i=\mathcal{N}(P_{i})\in \mathbb{R}^{N_{K} \times f},
    \end{aligned}
\end{equation}
where $\mathcal{N}(\cdot)$ denotes the KNN operation, and $N_{K}$ is the number of nearest neighbours. In contrast, the Farthest Point Sampling (FPS) algorithm
is employed to obtain the global key points as
\begin{equation}
    \begin{aligned}
    \overline{\mathbf{P}}_i=\mathcal{F}(P_{i})\in \mathbb{R}^{N_{F} \times f},
    \end{aligned}
\end{equation}
where $\mathcal{F}(\cdot)$ denotes the FPS operation with respective to $P_{i}$, and $N_{F}$ is the number of farthest points. Therefore, we can represent the features of $\underline{\mathbf{P}}_i$ and $\overline{\mathbf{P}}_i$ as
\begin{equation}
\begin{aligned}
    &\underline{\mathbf{X}}_i=\bm{E}_\alpha(\underline{\mathbf{P}}_i) \in \mathbb{R}^{N_K \times C},\\
    &\overline{\mathbf{X}}_i=\bm{E}_\alpha(\overline{\mathbf{P}}_i) \in \mathbb{R}^{N_F \times C},
\end{aligned}
\end{equation}
where $C$ denotes the feature dimension. 

In the subsequent multi-resolution attention mechanism, the query, key and value maps $\mathbf{Q}$, $\mathbf{K}$, and $\mathbf{V}$ of a point cloud is generated for $i$-th point as,
\begin{equation}
    \begin{aligned}
        \mathbf{Q}_i=&Net_{q}(\mathbf{X}_q^{i}) \in  \mathbb{R}^{1 \times C},\\
        \mathbf{K}_i=&Net_{k}(\overline{\mathbf{X}}_i \oplus \underline{\mathbf{X}}_i) \in  \mathbb{R}^{(N_{F}+N_{K}) \times C},\\
        \mathbf{V}_i=&Net_{v}(\overline{\mathbf{X}}_i \oplus \underline{\mathbf{X}}_i) \in  \mathbb{R}^{(N_{F}+N_{K}) \times C},\\
    \end{aligned}
\end{equation}
where $Net_q(\cdot)$, $Net_k(\cdot)$ , and $Net_v(\cdot)$ denotes the corresponding extractor mapping respectively, $\mathbf{X}_q^{i}$ is the $i$-th feature of the encoder output $\mathbf{X}_q$. Take single head as an example, the multi-resolution attention calculation process can be written as,
\begin{equation}
    \begin{aligned}
        &\mathbf{A}_i=\mathbf{Q}_i \otimes \mathbf{K}_i^{\top} \in \mathbb{R}^{1 \times (N_{F}+N_{K})},\\
        &\hat{\mathbf{Q}}_i=softmax(\mathbf{A}_i) \otimes \mathbf{V}_i^{\top} \in \mathbb{R}^{1 \times C},
    \end{aligned}
\end{equation}
where $\mathbf{A}_i$ is the $i$-th point attention map and $\hat{\mathbf{Q}}_i$ is the multi-resolution attention block output feature, $\hat{\mathbf{Q}} \in \mathbb{R}^{M \times C}$. The final feature of query point cloud can be obtained through the concatenation of $\hat{\mathbf{Q}}$ and ${\mathbf{X}_q}$,
\begin{equation}
    {\mathbf{X}}_{q}^{\prime}=\hat{\mathbf{Q}} \oplus {\mathbf{X}_q} \in \mathbb{R}^{M \times 2C}.
\end{equation}
Similarly, support feature $\mathbf{X}_{s}^{\prime}$
can be obtained in the corresponding way.

The support features might be dispersive due to a not well-trained encoder which would lead to a larger misclassification probability. Moreover, we would like to make support features far away from centers of other different classes. Hence, we adopt the center loss~\cite{qi2017contrastive} to guarantee the close distribution of the same class and sparse distribution of different classes in feature space.The loss function is:
\begin{equation}
\mathcal{L}_{r}=\frac{1}{2}\sum_{n=0}^{N}\sum_{i=1}^{M \times K} \frac{\left\|\mathbf{X}_{s,i}^{\prime,n}-C_{n}\right\|_2^2}{\left(\sum_{j=0, j \neq n}^{N}\left\|\mathbf{X}_{s,i}^{\prime,n}-C_j\right\|_2^2\right)+\eta}
\end{equation}
where $\mathbf{X}_{s,i}^{\prime,n}$ is the feature of the $i$-th support points of the $n$-th class, $\bm{C}_{n}$ is the center of class ${n}$ in feature space and $\eta$ is a hyper-parameter to avoid the denominator becoming 0, it is set to 1 as default. Finally, the compact center-regularized multi-prototype $\{\boldsymbol{\mu}^{n}\}_{n=0}^{N}$ can be generated using the standard method~\cite{prototypenet}.


\subsection{Label Network}

The label network serves to predict the class for each point in query point clouds based on the feature ${\mathbf{X}}_{q}^{\prime}$ and prototypes of support set $\{\boldsymbol{\mu}^{n}\}_{n=0}^{N}$. The basic idea of label propagation algorithm employed is to predict the label of unknown nodes from the information of labeled nodes using graph relationship between samples. For a $N$-way $K$-shot problem, the graph size is $
Z=N_p \times (N + 1) + |\mathcal{Q}| \times M$. $|\mathcal{Q}|$ represent the number of point clouds in the query set, $\mathbf{X}^{\prime} \in \mathbb{R}^{Z \times Z}$ denotes the feature matrix of all nodes. We additionally define feature matrix $\mathbf{W} \in \mathbb{R}^{Z \times Z}$, where its each value is calculated as
\begin{equation}
    \mathbf{W}_{ij}=\text{exp}(-\frac{\left\|\mathbf{{X}}_{i}^{\prime}- \underline{\mathbf{X}}_j^{\prime}\right\|_2^2}{2\sigma^2}),
\end{equation}
where ${\bm{X}}_{i}^{\prime}$ is the $i$-th feature in ${\mathbf{X}}$, and $\sigma$ is the standard deviation of the distance between two nodes. We follow~\cite{iscen_label_2019} to obtain the optimal prediction matrix $\mathbf{F}$ and the prediction output map $\mathbf{H}$,
\begin{equation}
    \begin{aligned}
    \mathbf{F}=(\mathbf{I}-\gamma (\mathbf{D}^{-1/2}&(\mathbf{W}+\mathbf{W}^{T})\mathbf{D}^{-1/2}))^{-1} \mathbf{L},\\
    \mathbf{H}_{i}^{n}=&\frac{\exp \left(\mathbf{F}_{i}^{n}\right)}{\sum_{j=0}^{N} \exp \left(\mathbf{F}_{i}^{j}\right)},
    \end{aligned}
\end{equation}
where $\mathbf{D}$ is a diagonal matrix whose value is the sum of corresponding rows of $\mathbf{W}$. $\gamma \in (0,1)$ is probability coefficient. $\mathbf{L} \in \mathbb{R}^{Z \times (N+1)}$ is a reference matrix, rows of $\mathbf{L}$ represent prototypes are one-hot labels and other represent unlabeled points(query points) are initial zero. 
The loss function of this block is a cross-entropy loss,
\begin{equation}
    \mathcal{L}_{l}=-\frac{1}{|\mathcal{Q}|} \frac{1}{M} \sum_{i=1}^{|\mathcal{Q}|} \sum_{j=1}^{M} \sum_{n=0}^{N} \delta\left(\bm{Y}_{j}^{i},n\right) \log \left(\mathbf{H}_{i, j}^{n}\right),
\end{equation}
where the ground truth label $\bm{Y}_{j}^{i}$ represents the $j$-th point in the $i$-th point cloud of query set and function $\delta(\cdot)$ is defined as:
\begin{equation}
    \delta(x,y)=\left\{
\begin{aligned}
1\quad & x=y \\
0\quad & x\ne y \\
\end{aligned}
\right.
\end{equation} 
Therefore, the complete loss function in the training phase of our framework is
\begin{equation}
  \mathcal{L}=\mathcal{L}_{l}+\lambda\mathcal{L}_{r},
  \label{eq:loss}
\end{equation}
where $\lambda$ is the coefficient to adjust the intensity of centering and concentration.
\begin{figure}[!t]
    \centering
    \includegraphics[width=8.5cm]{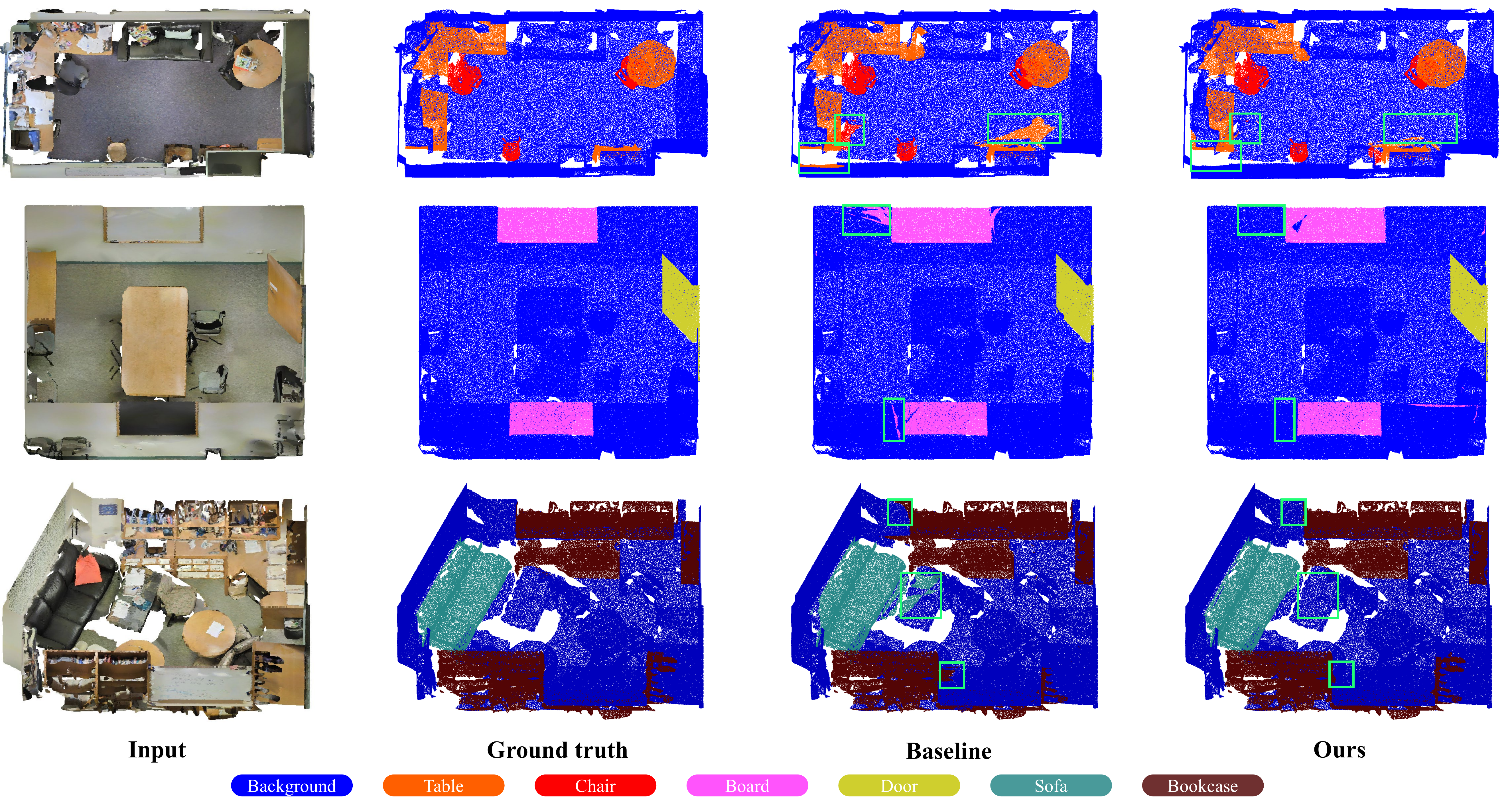}
    \caption{Result visualization and comparison on s3dis segmentation task.}
    \label{fig:res}
\end{figure}

\section{EXPERIMENTS}

\subsection{Implementation Details}

The network structure is shown in Fig. \ref{fig:netstructure}. In the pretrain phase, S3DIS~\cite{s3dis} and ShapeNet~\cite{shapenet} are adopted to figure out the influence of pretraining on different datasets, we simply add an MLP network as our classifier for prediction. Our batch size is set to 16, the initial learning rate is set to 0.001 and the pretrain epochs are set to 120. The augmentor is trained in advance on ShapeNet dataset. The $k$-NN number for the DGCNN-based encoder is set to 200. In the few-shot learning phase, the depth of the MRA block is set to 1 for computing efficiency, we additionally set the centering intensity coefficient $\lambda$ as 0.1 the label propagation parameter $\gamma$ is set to 0.9 to ensure the performance. Moreover, the $k$-NN number and fps ratio in the MRA block is set to 250 and 0.4, respectively. After initializing the feature extractor weights with the pretrained encoder, we set the Adam optimizer with an initial learning rate of 0.0005 for the feature extractor module, for other modules the learning rate is set to 0.001. Two subsets $S^{0}$ and $S^{1}$ are split for test in standard way~\cite{zhao_few-shot_2021}, we train our model with 100 episodes and 40,000 iterations on both subsets.

\subsection{Baselines}
For baselines, we choose attMPTI~\cite{zhao_few-shot_2021} and typical fine-tuning network~\cite{shaban2017one} as our comparative models. AttMPTI is a simple but effective network, it extracts prototypes from support feature and uses vanilla label propagation to predict query samples. For fine-tuning network, we freeze the weights of the pretrained feature extractor for the fine-tuning network and only update the parameters in the MLP-based segmentor of it to avoid overfitting.  In this work, we only use xyzXYZ without RGB in all datasets for all the experiments conducted in this work.

\begin{table}[t]
    \centering
    \caption{\textbf{Comparison of Methods on s3dis Dataset (\%)}}
    \begin{tabular}{c c c c c c c}
        \toprule
        \multirow{2}{*}{Method} & \multicolumn{3}{c}{2-Way 1-Shot} & \multicolumn{3}{c}{2-Way 5-Shot} \\
        \cmidrule(l){2-4} \cmidrule(l){5-7}
        & $S^{0}$ & $S^{1}$ &mean & $S^{0}$ & $S^{1}$ &mean\\
        \midrule
        Fine-Tuning & 36.04&36.69&37.37 &55.19&55.40&55.30  \\ 
        attMPTI & 52.29&53.84&53.07 &61.33&62.42&61.88 \\
        \makecell{Ours+VWA}& 54.48 &54.02 &54.26&62.16&61.89&62.03\\
        Ours    &  \textbf{54.80}& \textbf{54.13} &\textbf{54.47}&\textbf{63.26}&\textbf{62.61}&\textbf{62.84}  \\
    \bottomrule
    \end{tabular}
    \label{tb:rescomp}
\end{table}
\subsection{Comparison Result}
Table \ref{tb:rescomp} shows the result comparison between our framework and baselines. It shows that our model outperforms the baselines in all different few-shot settings, the cost is that our framework trains a little bit slower than the fine-tuning network and attMPTI due to feature concatenation. We also test vanilla window-attention from Stratified Transformer in Table~\ref{tb:rescomp}, denoted as ours+VWA, which shows it is lower than our approach. This agrees with our assumption that the window-attention may lose some global information. The fine-tuning network has an unsatisfied performance especially on 2 way 1-shot task, this is because the embedding features from the feature extractor is insufficient to train the segmentor of the fine-tuning network, and with the scarcity of data, the segmentor hardly can learn the features well. This result also indexes the importance of the attention mechanism which helps in learning a more representative embedding space. The vanilla attention block might projects redundant attention features between query and background points to the feature space.

\subsection{Ablation Study}
We conduct an ablation studies to verify the effectiveness of each component in our method, table~\ref{tb:ablres} illustrated the result of our experiments. We make the evaluation on the S3DIS $S^{0}$ dataset with the 2-way 1-shot few-shot setting. The verification was made by removing each component from the final model, \textit{i.e.}, Exp.~$\text{\uppercase\expandafter{\romannumeral8}}.$
Compared to Exp.~$\text{\uppercase\expandafter{\romannumeral1}}$, Exp.~$\text{\uppercase\expandafter{\romannumeral2}}$ shows the improvement of contrastive pretrain. Obviously, it enhances the segmentation performance. The result in Exp.~$\text{\uppercase\expandafter{\romannumeral3}}$ and Exp.~$\text{\uppercase\expandafter{\romannumeral4}}$ illustrates that the performance of using the centering-based multi-prototype or the MRA block individually is lower than employing the contrastive pretrain. This comparison shows the significance of a well-pretrained encoder in few-shot semantic segmentation.

Table \ref{tb:contrastpretrain} shows the effectiveness of our contrastive self-supervision pretrain. Without any other additional operations, we only adopt our contrastive self-supervised pretrain to attMPTI in the 2-way 1-shot few-shot semantic segmentation, the result shows that it significantly improves the few-shot segmentation performance. Particularly, Table~\ref{tb:contrastpretrain} indicates that it greatly helps to surmount the overfitting issue in small-dataset pretrain situation. As a result, our contrastive self-supervised pretrain is a general plug-and-play module, which is compatible to many other point cloud few-shot semantic segmentation frameworks.

\subsection{Case Study on Practical CAD/CAM Segmentation}
\begin{figure}[!t]
  \centering
  \includegraphics[scale=0.24]{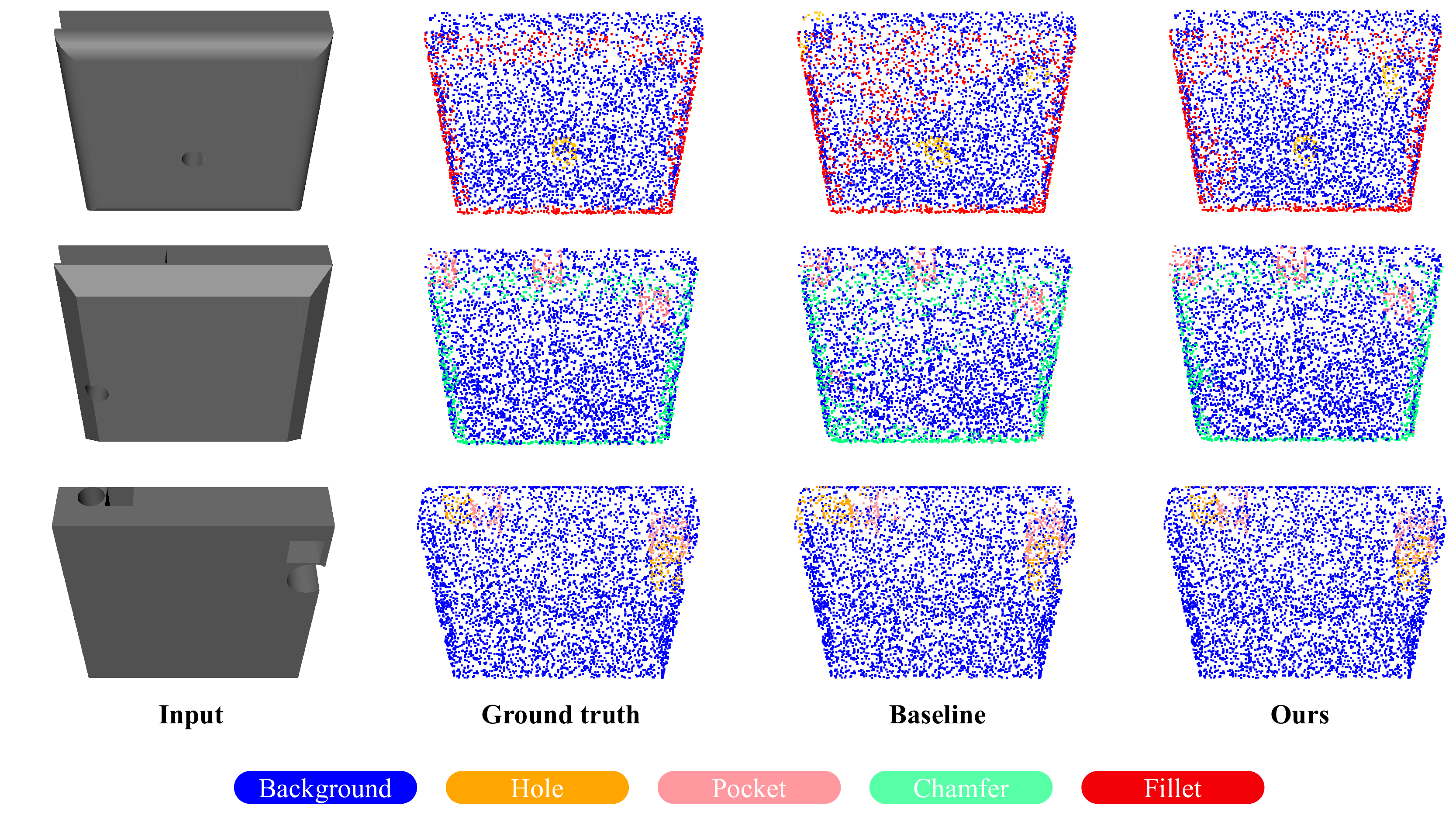}
  \caption{Result visualization and comparison on 3D CAD/CAM segmentation task}
  \label{fig:pcr}
\end{figure}

To demonstrate the practical effectiveness of our approach on real-world small-data tasks, we prepare a few-shot dataset from EIF 3D CAD ~\cite{lee2022dataset} and evaluate our algorithm on it, which is an important step in advanced automation workflow especially in CAD/CAM process. The dataset, named as  phcf, is a CAD/CAM 3D semantic segmentation task, which includes four types of features, \textit{i.e.} pocket, hole, chamfer, and fillet. However, due to the similar shape between point clouds, this dataset is easily overfitted in the training due to the lack of complexity. Table~\ref{tb:pcrescomp} illustrated the best and mean results of different methods, repeated 5 times, and the visualization is shown in Fig.~\ref{fig:pcr}. From the results, it can see that our approach is more effective in dealing with real-world data-scarce tasks, where those dataset are often small-scale and will be easily-overfitted in the supervision. Notably, Table~\ref{tb:contrastpretrain} also shows that without an additional large-scale dataset, our approach improves significantly for only contrastive pretrain using the same-domain dataset (phcf). Since our self-supervision does not require annotated dataset, it greatly facilitates the real-world few-shot tasks to conduct the effective pretrain on the close-domain dataset, which finally improves the segmentation accuracy.  

\begin{table}[ht]
    \centering
    \caption{\textbf{2-Way 1-Shot Ablation Study on S3DIS Dataset ($S^{0}$)}} 
    \scalebox{0.9}{\begin{tabular}{c c c c c}
      \toprule
     ID & \makecell{Multi-Resolution\\ Attention} & \makecell{Centering-Based\\ Multi-Prototype} & \makecell{Contrastive\\Pretrain} & \makecell{Accuracy\\(mIoU)} \\
      \midrule
    \uppercase\expandafter{\romannumeral1} & & & &52.29\\
    \uppercase\expandafter{\romannumeral2} & & &\checkmark&53.78\\
    \uppercase\expandafter{\romannumeral3} & &\checkmark& &52.56\\
    \uppercase\expandafter{\romannumeral4} &\checkmark& & &52.92\\
    \uppercase\expandafter{\romannumeral5} &\checkmark&\checkmark& &53.46\\
    \uppercase\expandafter{\romannumeral6} &\checkmark& &\checkmark&54.23\\
    \uppercase\expandafter{\romannumeral7} & &\checkmark&\checkmark&54.07\\
    \uppercase\expandafter{\romannumeral8} &\checkmark&\checkmark&\checkmark&\textbf{54.80}\\
      \bottomrule
    \end{tabular}}
    \label{tb:ablres}
\end{table}

\begin{table}[ht]
    \centering
    \caption{\textbf{Comparison of w/wo Contrastive Pretrain}} 
    \scalebox{0.9}{\begin{tabular}{c c c c c}
      \toprule
    \makecell{Pretrain\\ Dataset} & \makecell{Train\\ Dataset} & \makecell{Supervised\\Pretrain Result} & \makecell{Contrastive\\Pretrain Result} & $\Delta$\\
      \midrule
    S3DIS&S3DIS &52.29&53.78 & +1.49\\
    ShapeNet&S3DIS &52.87&54.21&+1.34 \\
    phcf&S3DIS&45.15&50.19& +5.04\\
    S3DIS&phcf &49.40&55.38& +5.98\\
    ShapeNet&phcf &52.34&56.80& +4.46\\
    phcf&phcf &29.97&47.11& \textbf{+17.14}\\
      \bottomrule
    \end{tabular}}
    \label{tb:contrastpretrain}
\end{table}

\begin{table}[!ht]
    \centering
    \caption{\textbf{Comparison Methods on phcf Dataset (\%)}}
    \scalebox{0.9}{\begin{tabular}{c c c c c c c}
        \toprule
        \multirow{2}{*}{Method} & \multicolumn{2}{c}{2-Way 1-Shot} & \multicolumn{2}{c}{2-Way 3-Shot} & \multicolumn{2}{c}{2-Way 5-Shot}\\
        \cmidrule(l){2-3} \cmidrule(l){4-5} \cmidrule(l){6-7}
        & Best & Mean & Best & Mean & Best & Mean\\
        \midrule
        Fine-Tuning & 41.54 & 39.01&46.39  &46.32&50.41 &49.71   \\ 
        attMPTI & 49.23&48.64&51.81 &51.59&53.02&52.49  \\
        Ours    &  \textbf{52.71}& \textbf{51.92} &\textbf{53.70}&\textbf{52.83}&\textbf{54.24}&\textbf{54.06}  \\
    \bottomrule
    \end{tabular}}
    \label{tb:pcrescomp}
\end{table}

\section{CONCLUSIONS}
In this work, we propose and develop a noteworthy few-shot point cloud semantic segmentation framework.Firstly, by using contrastive self-supervised pretrain, our approach is capable to reduce the feature bias and overfitting due to the class-specific fully-supervision. Moreover, it enables to use unlabeled datasets for close-domain pretrain in real-world tasks. Secondly, the MRA mechanism helps to mitigate the influence of redundant background information without overlooking local or global features. Thirdly, an intra-class centering multi-prototype generation is adopted to improve the accuracy of prediction. Comprehensive experiments shows that our approach achieves better accuracy comparing to the existing methods. Finally, a practical case study is conducted to shows the effectiveness of our framework in a real-world application.

\section*{ACKNOWLEDGMENT}
This research is supported by the National University of Singapore under the NUS College of Design and Engineering Industry-focused Ring-Fenced PhD Scholarship programme. This research is also supported by A*STAR under its ``RIE2025 IAF-PP Advanced ROS2-native Platform Technologies for Cross sectorial Robotics Adoption (M21K1a0104)" programme. The authors would like to acknowledge useful discussions with Dr.Arno Zinke from Hexagon, Manufacturing Intelligence Division, Simufact Engineering GmbH.


\end{document}